\documentclass[conference]{IEEEtran}
\IEEEoverridecommandlockouts
\usepackage{diagbox}
\usepackage{cite}
\usepackage{amsmath,amssymb,amsfonts}
\usepackage{graphicx}
\usepackage{textcomp}
\usepackage{xcolor}
\usepackage{tabularx}
\usepackage{times}
\usepackage{epsfig}
\usepackage{graphicx}
\usepackage{amsmath}
\usepackage{amssymb}
\usepackage{textcomp}
\usepackage{subcaption}
\usepackage{balance}
\usepackage[labelfont=bf]{caption}
\usepackage{multirow}
\usepackage{array}

\newcolumntype{P}[1]{>{\centering\arraybackslash}p{#1}}
\usepackage{notoccite}

\usepackage{algorithm,algpseudocode}
\usepackage{amsmath}
\usepackage{amsfonts}
\usepackage{amssymb}

\usepackage{lipsum}
\usepackage{multicol}
\newcommand\blfootnote[1]{%
  \begingroup
  \renewcommand\thefootnote{}\footnote{#1}%
  \addtocounter{footnote}{-1}%
  \endgroup
}

\def\BibTeX{{\rm B\kern-.05em{\sc i\kern-.025em b}\kern-.08em
    T\kern-.1667em\lower.7ex\hbox{E}\kern-.125emX}}

\title{An AI-powered Smart Routing Solution for Payment Systems}

\makeatletter
\newcommand{\linebreakand}{%
  \end{@IEEEauthorhalign}
  \hfill\mbox{}\par
  \mbox{}\hfill\begin{@IEEEauthorhalign}
}
\makeatother

\author{
  \IEEEauthorblockN{Ramya Bygari\IEEEauthorrefmark{1}}
  \IEEEauthorblockA{
    \textit{Razorpay }\\
    Bengaluru, India \\
    ramya.b@razorpay.com}
  \and
  \IEEEauthorblockN{Aayush Gupta\IEEEauthorrefmark{1}}
  \IEEEauthorblockA{
    \textit{Razorpay }\\
    Bengaluru, India \\
    aayush.gupta@razorpay.com}
  \and
  \IEEEauthorblockN{Shashwat Raghuvanshi\IEEEauthorrefmark{1}}
  \IEEEauthorblockA{
    \textit{Razorpay }\\
    Bengaluru, India \\
    shashwat.raghuvanshi@razorpay.com}
  \linebreakand 
  \IEEEauthorblockN{Aakanksha Bapna}
  \IEEEauthorblockA{
    \textit{Razorpay }\\
    Bengaluru, India \\
    aakanksha.bapna@razorpay.com}
  \and
  \IEEEauthorblockN{Birendra Sahu}
  \IEEEauthorblockA{
    \textit{Razorpay }\\
    Bengaluru, India \\
    birendra.sahu1973@gmail.com}
}

\begin{document}

\maketitle
\blfootnote{\IEEEauthorrefmark{1}Authors contributed equally}
\begin{abstract}
In the current era of digitization, online payment systems are attracting considerable interest. 
Improving the efficiency of a payment system is important since it has a substantial impact on revenues for businesses. A gateway is an integral component of a payment system through which every transaction is routed. In an online payment system, payment processors integrate with these gateways by means of various configurations such as pricing, methods, risk checks, etc. These configurations are called terminals. Each gateway can have multiple terminals associated with it. Routing a payment transaction through the best terminal is crucial to increase the probability of a payment transaction being successful. Machine learning (ML) and artificial intelligence (AI) techniques can be used to accurately predict the best terminals based on their previous performance and various payment-related attributes. We have devised a pipeline consisting of static and dynamic modules. The static module does the initial filtering of the terminals using static rules and a logistic regression model that predicts gateway downtimes. Subsequently, the dynamic module computes a lot of novel features based on success rate, payment attributes, time lag, etc. to model the terminal behaviour accurately. These features are updated using an adaptive time decay rate algorithm in real-time using a feedback loop and passed to a random forest classifier to predict the success probabilities for every terminal. This pipeline is currently in production at Razorpay routing millions of transactions through it in real-time and has given a 4-6\% improvement in success rate across all payment methods (credit card, debit card, UPI, net banking). This has made our payment system more resilient to performance drops, which has improved the user experience, instilled more trust in the merchants, and boosted the revenue of the business. 
\end{abstract}


\section{Introduction}

\begin{figure}[t]
\begin{center}
   \includegraphics[width=1.0\linewidth]{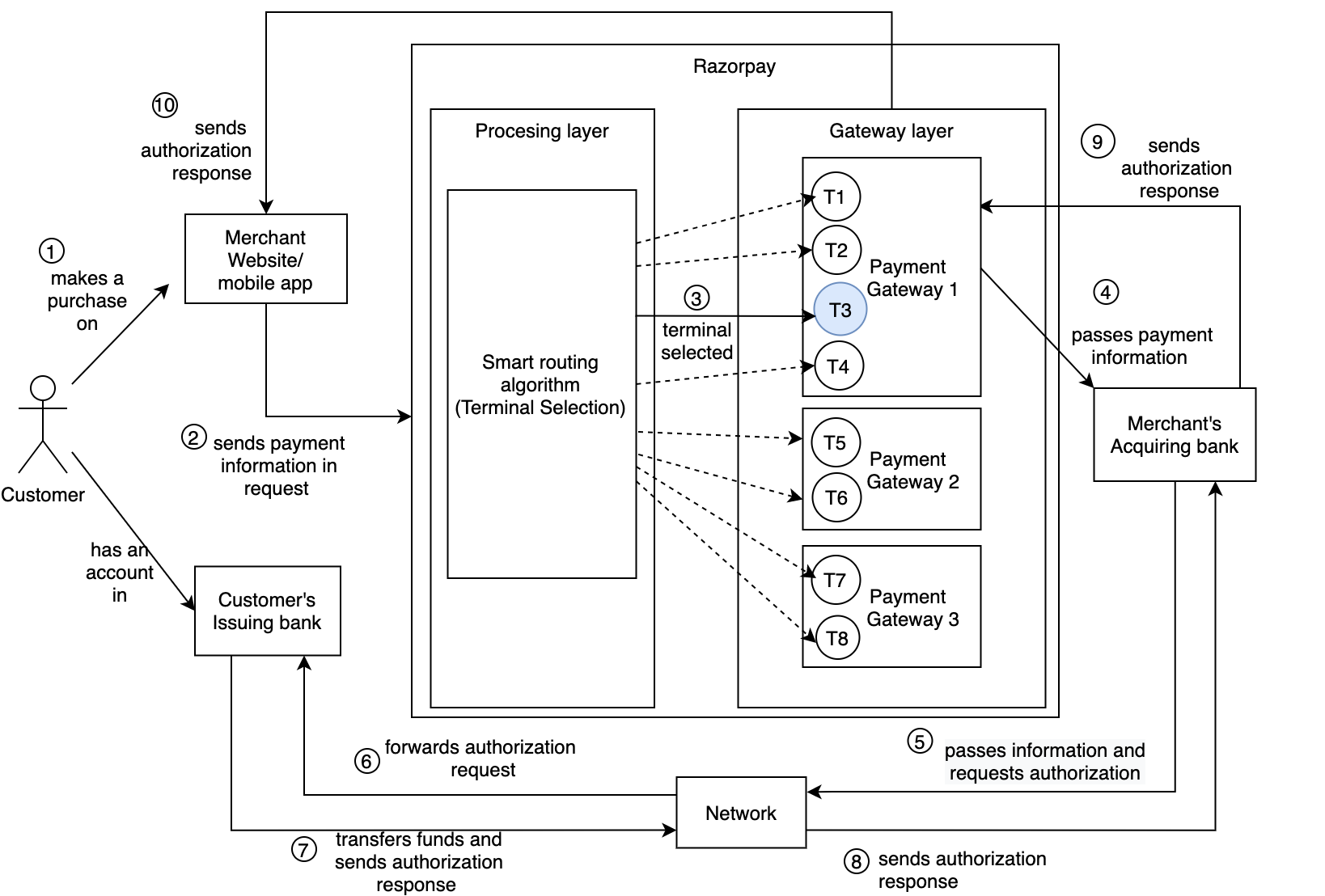}
\end{center}
   \caption{The life-cycle of a typical payment transaction. 
   The Smart Routing System selects the best terminal for a payment.}
   \label{Payment Flow}
\end{figure}

Razorpay is an online payment gateway aggregator that facilitates the ease of initiating and accepting payments without the hassle of dealing with banks. It acts as a bridge between individuals/merchants and banking institutions.  The basic components of a typical payment flow are as follows:
\begin{itemize}
 \item Customer: Any individual who initiates an online payment.
\item Issuing bank: The bank that processes the online payments of the customer.

\item Merchant: Any business that sells goods or services online. Usually, this is a website or mobile application where customers shop.
\item Acquiring bank: The bank that processes the online payments of a merchant.
\item Processor: The companies such as Razorpay that provide a single point of contact for merchants and several third-party banks to process online payment transactions.
\item Payment Gateway: The technology that processes online payments. This is usually owned by a bank, in most cases, the acquiring bank. Every bank has its own payment gateway and these vary in success rates to process an online payment.
\item Terminal: An entity used to store configuration parameters (along with some metadata) to successfully communicate with a gateway.
\item Network: The companies such as Visa or NPCI that connect the issuing bank and the acquiring bank to facilitate online payments.
\end{itemize}

The life cycle of an online payment is illustrated in Fig. \ref{Payment Flow}. The components described earlier interact with each other in a sequence of steps as described below:
\begin{enumerate}
 \item A typical online payment flow starts with a customer initiating a payment on any merchant website/app. 
This creates a payment request containing all the information necessary for payment (E.g. amount, requester, receiver, token) which is sent to the payment processor (here Razorpay).

 \item This payment request arrives at the processing layer, which uses the Smart Routing Algorithm to route it to the most appropriate terminal from the available payment gateways.
\item The payment information is passed to the acquiring bank from the payment gateway of the best terminal (selected above) to initiate the transaction.
\item The merchant’s bank requests authorization for the payment from the network.
\item The network forwards the authorization request to the customer’s issuing bank.
\item The issuing bank authorizes the payment, debits the funds from the customer’s bank and then sends the authorization response back to the network.
\item The network then sends the authorization response to the merchant’s acquiring bank and credits the funds into the bank.
\item The authorization response is then sent to the payment processor.
\item The payment processor then sends the confirmation of payment to the merchant’s website/app.
\end{enumerate}

Any payment transaction results in either a success or a failure. Payment failures can be categorised broadly into 2 categories: customer-related failures, non-customer-related failures. Customer-related failures consist of wrong one-time password (OTP), wrong Card Verification Value (CVV), payment timeouts, etc which are unavoidable. Non-customer-related failures occur at the gateways due to performance degradation due to the following reasons:
(i) The Gateway may be overloaded with more capacity than it can handle, which leads to a sudden decrease in success rates.
(ii) The Bank server system may go down or may go under maintenance, leading to the complete failure of the gateway.

This implies that gateway selection for any payment transaction plays a crucial role in preventing failure. Hence, to maximize the success rate for payments, the Smart Routing algorithm is used. It utilises hundreds of parameters in real-time to identify the best performing terminal.
Moreover, the Smart Routing algorithms may also be used to fulfil certain other business requirements like pricing optimization or merchant-tailored gateway routing.

When the payment system receives a request, the Smart Routing system analyses all available routes and returns a list of terminals sorted by their probabilities of success.  This system provides the payment gateway interface with the best possible option to transact by considering multiple factors such as:
(i) Historical data for success rate, downtime, gross merchandise value (GMV), ticket size, etc.
(ii) Volume across different entities like a merchant, card, bank, customer, etc.
(iii) Pricing data for various dimensions like card, net banking, terminal, merchant, exclusivity, gateway, etc.

\begin{figure*}[t]
\begin{center}
   \includegraphics[width=1.0\linewidth]{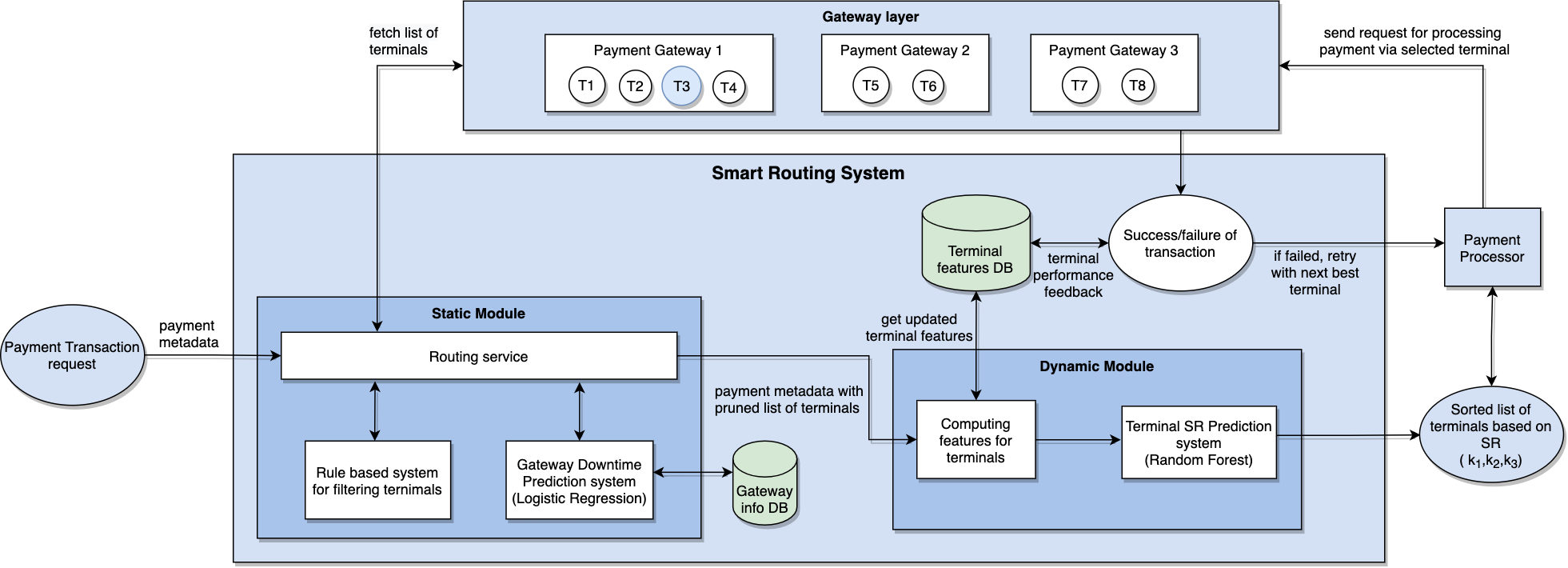}
\end{center}
   \caption{Architecture of the Smart Routing System.}
   \label{Proposed method}
\end{figure*}

\section {Related work}
Smart Routing is a common problem occurring in payments that can optimize the journey of a payment resulting in the growth of business \cite{wiley_book}. This has been solved by many companies like PayCore.io Limited \cite{paycore}, Processout Limited \cite{processout}, Spreedly Inc. \cite{spreedly}, etc.

To the best of our knowledge, PayU's research \cite{trivedi2018stochastic} is the only work similar to this paper that uses artificial intelligence for routing payments through appropriate terminals. They have devised an approach that employs reinforcement learning by using the success and failure of transactions as the rewards and punishments in their algorithm. Based on these values, they are maximising the expected rewards of a transaction over a period of time.
PayU’s Smart Routing system is good at capturing the features associated with the terminal performance based on time, however, it suffers from a few disadvantages:
\begin{itemize}
\item
 As per the information presented in their paper, it characterizes the behaviour of a terminal only based on its past performance. However, there are many other attributes of a transaction like method of transaction, amount, IP, merchant, etc. which can help in modeling the terminal behaviour better.
\item In general, RL systems are Markovian \cite{sutton2018reinforcement} in nature, however, the nature of a payment terminal may not be strictly Markovian.

\end{itemize}

Considering these factors, we have devised an approach that uses tree-based supervised machine learning algorithms that predicts the most suitable terminal for routing the payment using the current and historical data available for a transaction.
\section{Overview}

The Smart Routing pipeline consists of two main modules: a static module and a dynamic module. The architecture of the proposed method is shown in Fig. \ref{Proposed method}.
The payment system receives a transaction request from a merchant portal. It arrives at the \textbf{static module} for relevant terminal selection. The routing service fetches the list of available terminals from the gateway layer and applies merchant/business based rules to filter out irrelevant terminals. The gateway downtime prediction module uses past gateway information to predict gateway downtimes and filter out the terminals corresponding to that gateway. The filtered input terminals are sent to the \textbf{dynamic module} for success probability assignment for each terminal. In the dynamic module, various features like time-window-based features, event-based features, etc. are computed from the past data and saved in the database. For a new payment request, the feature values are fetched and decayed at the current timestamp and probability predictions are made for each terminal. This module returns a list of terminals sorted in the order of their success probabilities. The payment is initiated through the terminal with the highest probability. If this payment fails, it is routed to the next best terminal. The feature values for the terminal that processed the payment are sent along with the outcome as feedback to update the features for the terminal in the database. The creation and transformation of the features that serve as an input to the static system (Section \ref{static system}) and the dynamic system (Section \ref{Dynamic System}) is explained in Section \ref{Feature Creation}. The real-time feature update occurs through a feedback loop (Section \ref{feedback loop}) using an adaptive decay-rate algorithm (Section \ref{feature decay}).

\section{Problem Definition}
Each payment transaction has two possible outcomes, success or failure. As shown in Fig. \ref{Proposed method}, when the Smart Routing system receives a payment transaction request, it needs to identify the most suitable terminal to increase the success probability of the transaction. The Smart Routing system takes a terminal list as input and assigns a success probability to each terminal based on its past transactions and its current performance. Then, the payment is processed through the terminal with the highest predicted probability of success.
\par



For each transaction request, let the set of terminals be  $k$, where  $k_i$, $ i \in [1, n] $ represents each terminal.

Let the random forest classifier be an ensemble model of $x$ trees each represented as $t_r$, $ r \in [1, x] $ used to calculate the probability of success for each  $k_i$, $ i \in [1, n] $. The probability of success for each terminal is given by the count of trees that predicted probability greater than $0.5$. The payment transaction system
returns the list of terminals sorted on the basis of their success probabilities as shown in Algorithm  \ref{Algorithm1}.







\algnewcommand\algorithmicforeach{\textbf{for each:}}
\algnewcommand\ForEach{\item[ \algorithmicforeach]}

\begin{algorithm}[H] 
\caption{Prediction Algorithm}\label{Algorithm1}
\begin{algorithmic}[1]
\Require{ $k$, list of $n$ terminals to be sorted }


\State $k\_dict$ = $dict()$ \# to store success probabilities 

\State \textbf{for each} $k\_i$ in $k$:
\State \indent \# probability calculation from random forest
\State \indent $k\_dict[k\_i]=({\sum_{r=1}^{r=x}(Prob(t_r)>0.5)})/{x}$
\State \textbf{end for}
\State $k\_sorted = desc\_sort\_by\_values(k\_dict).keys()$

\Ensure{ $k\_sorted$, list of terminals sorted by their success probabilities}
\end{algorithmic}
\end{algorithm}


\section{Methodology}
\label{methodology}

The Smart Routing pipeline consists of two main components: the static module and the dynamic module. The static module tracks the real-time performance of the gateway and filters the input terminal list on the basis of business agreements with various merchants and banks. The dynamic module predicts the success probabilities for each terminal. The terminal with the maximum success probability is chosen as the terminal through which the transaction should occur. Based on the payment outcome, the feature values corresponding to the terminal are updated using an adaptive time-decay algorithm to assign more weight to the recent payment outcomes over the past payment outcomes. Each of these components is further explained in detail in the subsequent sections.

\subsection{Feature Creation and Transformation}
\label{Feature Creation}
A terminal’s performance is measured by its success rate (SR). The success rate of a terminal $k$ during a specific time frame, $[T\textsubscript{1}, T\textsubscript{2}]$ is defined as the ratio of the number of successful transactions to the total number of transactions that occurred through the terminal $k$ between $[T\textsubscript{1}, T\textsubscript{2}]$
as shown in equation \ref{SR}.

\begin{equation} 
\label{SR}
\begin{split}
\textup{SR} = \frac{\text{(Number of successful transactions)\textsubscript{(k, [T\textsubscript{1},T\textsubscript{2}])}}}{\text{\text{}(Total number of transactions)\textsubscript{(k, [T\textsubscript{1},T\textsubscript{2}])} }}
\end{split}
\end{equation}
 
Hence, it is evident that the outcome (i.e, success or failure) of the transactions through a terminal determines the success rate of the terminal, which in turn aids in establishing the performance of that terminal. The performance of the terminal can be efficiently tracked by creating three types of features that can effectively incorporate periodical and seasonal changes:
\begin{enumerate}
    \item Time-window-based features 
    \item Event-based features
    \item Overall features
\end{enumerate}

\begin{table}[]
\begin{center}
\caption{Calculation of features at different levels.}    

\renewcommand{\arraystretch}{1.2}
\centering
\begin{tabular}{|p{24mm}|P{16mm}|P{16mm}|P{16mm}|}
\hline
\textbf{Types of Features}   & \textbf{Terminal Level} & \textbf{Gateway Level} & \textbf{Payment System Level} \\ \hline
Time-window-based Features & Yes            & Yes           & Yes                  \\ \hline
Event-based Features & Yes            & Yes           & Yes                  \\ \hline
Overall Features    & No             & No            & Yes                  \\ \hline

\end{tabular}

\label{heirarchyfeatures}
\end{center}
\end{table}

\par
Time-window-based and event-based features are calculated for the terminals individually, whereas the overall features are calculated for the payment system collectively. Moreover, time-window-based and event-based features for a gateway, say $G\textsubscript{i}$, are calculated by aggregating the feature values of all the terminals that map to $G\textsubscript{i}$. Table \ref{heirarchyfeatures} represents the different levels (terminal/gateway/payment system) at which these features are calculated.

\subsubsection{Time-window-based features}
Time-window-based feature values for a terminal are calculated based on the outcome of the transactions in the last $t$ seconds through it ($t$ takes any positive integer value. Choosing optimal values for $t$ which closely represent the real-time terminal features is explained in Section \ref{feature selection}). If the current timestamp of the transaction is $T$, the corresponding time frame for which the features for a terminal are calculated is $[T-t, T]$, i.e, transactions between $[T-t, T]$ are considered for feature calculation.  For example,  consider the inherent characteristics representing a terminal $k$ to be $f\textsubscript{1}, f\textsubscript{2}, f\textsubscript{3}, .. f\textsubscript{n}$. 
and let $F$ denote all possible combinations of $f\textsubscript{1}, f\textsubscript{2}, f\textsubscript{3}, .. f\textsubscript{n}$. 
Creating $t$ $seconds$ time-window-based features for each element in $F$ implies calculating the success rate values (using equation \ref{SR}) for the time frame $[T-t,T]$. Table \ref{features} illustrates an example of a terminal with $f\textsubscript{1}, f\textsubscript{2}, f\textsubscript{3}$ characteristics. $Yes$ in the cell implies that the particular characteristic (column name) of a terminal is considered in the creation of the corresponding feature (row name), and this feature can further be used to track the terminal’s performance. $No$ in the cell implies that the particular characteristic of a terminal is not considered in the creation of the corresponding feature.  
For instance, feature values at current timestamp $T$ for $f\textsubscript{1}\_f\textsubscript{2}\_5s$ is equivalent to considering all the transactions that are processed through terminals with features $f\textsubscript{1}$ and $f\textsubscript{2}$ in the last $5$ seconds, and each transaction is given a weight according to its recency. The basic concept is to decrement the contribution of older values by multiplying them with weights that are inversely proportional to the age of the event. Assigning equal weight to all the past transactions independent of their time does not capture recency, however the decayed features on the other hand capture recency by taking a time factor into consideration. This encapsulates the ongoing performance of the terminal. This has been further explained in Section \ref{feature decay}.
For the 1st payment transaction, the values corresponding to the time-window-based features are equal to $1$. Table \ref{time based} illustrates an example of calculation of feature  $f\textsubscript{1}\_30s$, where $30s$ (half-life for the feature) and the payment timestamp are used for decaying the features.
The outcome column in the Table \ref{time based} represents the payment outcomes where $1$ indicates payment success and $0$ indicates payment failure. The SR column is calculated as an expanding mean over the payment outcomes. This table shows that the feature $f\textsubscript{1}\_30s$ where time is taken into consideration represents recency more accurately than the SR where time is not taken into consideration.
\begin{table}[]
\begin{center}
\caption{Characteristics of the terminal to be considered for corresponding feature creation.}
\centering
\begin{tabular}[3]{|p{40mm}|P{12mm}|P{9mm}|P{9mm}|}

\hline
\backslashbox[40mm]{\textbf{Features (F)}}{\textbf{Characteristics (f)}}  & \textbf{f1}  & \textbf{f2}  & \textbf{f3}  \\ \hline
f\textsubscript{1}\_t         & Yes & No  & No  \\ \hline
f\textsubscript{2}\_t         & No  & Yes & No  \\ \hline
f\textsubscript{3}\_t         & No  & No  & Yes \\ \hline
f\textsubscript{1}\_f\textsubscript{2}\_t     & Yes & Yes & No  \\ \hline
f\textsubscript{2}\_f\textsubscript{3}\_t     & No  & Yes & Yes \\ \hline
f\textsubscript{1}\_f\textsubscript{3}\_t     & Yes & No  & Yes \\ \hline
f\textsubscript{1}\_f\textsubscript{2}\_f\textsubscript{3}\_t & Yes & Yes & Yes \\ \hline
\end{tabular}

\label{features}
\end{center}
\end{table}

\begin{table}[]
\begin{center}
\caption{Example of time-window-based feature calculation (t=30s).}
\renewcommand{\arraystretch}{1.2}
\centering
\begin{tabular}{|p{13mm}|P{12mm}|P{14mm}|P{14mm}|P{12mm}|}

\hline
\textbf{Payment} & \textbf{Outcome} & \textbf{Timestamp}    & \textbf{SR} & \textbf{f\textsubscript{1}\_30s}    \\ \hline
1       & 1  & 1629829802   & 1/1 = 1.00& 1.0000       \\ \hline
2       & 0  & 1629829803   & 1/2 = 0.50& 0.6402  \\ \hline
3       & 1  & 1629829804   & 2/3 = 0.67 & 0.5869  \\ \hline
4       & 1  & 1629829805   & 3/4 = 0.75&0.4998 \\ \hline
5       & 0  & 1629829806   & 3/5 = 0.60& 0.5234  \\ \hline
6       & 1  & 1629829807   & 4/6 = 0.67& 0.5568   \\ \hline
7       & 1  & 1629829808   & 5/7 = 0.71 & 0.5937  \\ \hline
\end{tabular}

\label{time based}
\end{center}
\end{table}

\subsubsection{Event-based features}
Event-based features at a particular timestamp $T$, take into consideration the outcome of the transactions through a terminal $k$ in the past $e$ events where $e$ takes any positive integer value. (Choosing the values which closely represent the real-time terminal features is explained in Section \ref{feature selection}). For example,  $f\textsubscript{1}\_f\textsubscript{2}\_10e$ represents the success probability of the past $10$ successive transactions that went through terminals with characteristics $f\textsubscript{1}$ and $f\textsubscript{2}$. This success probability is calculated as a sliding mean of $e$ window length over the transactions. If at a particular timestamp there exists less than $e$ events/transactions prior, then the success probability is a function of their expanding mean. For the 1st payment transaction, the values corresponding to the event-based features are equal to $1$.  Table \ref{event-based} illustrates an example of $f\textsubscript{1}\_5e$ and $f\textsubscript{1}\_10e$ calculation. The outcome column in the Table \ref{event-based} represents the payment outcomes where $1$ indicates payment success and $0$ indicates payment failure.

\begin{table}[]
\begin{center}
 \caption{Example of event-based feature calculations.}   
 \centering
\begin{tabular}{|p{16mm}|P{16mm}|P{16mm}|P{16mm}|}

\hline
\textbf{Payment} &  \textbf{Outcome} & \textbf{f\textsubscript{1}\_5e}   & \textbf{f\textsubscript{1}\_10e}   \\ \hline
1       & 1       & 1.000    & 1.000     \\ \hline
2       & 0       & 0.500  & 0.500   \\ \hline
3       & 1       & 0.667 & 0.667  \\ \hline
4       & 0       & 0.500  & 0.500   \\ \hline
5       & 1       & 0.600  & 0.600   \\ \hline
6       & 0       & 0.400  & 0.500   \\ \hline
7       & 1       & 0.600  & 0.571  \\ \hline
8       & 1       & 0.600  & 0.625 \\ \hline
9       & 1       & 0.800  & 0.667  \\ \hline
10      & 0       & 0.600  & 0.600   \\ \hline
11      & 1       & 0.800  & 0.600   \\ \hline
\end{tabular}

\label{event-based}
\end{center}
\end{table}
\subsubsection{Overall features}
The overall features are calculated for every transaction, not taking into consideration the characteristics of the terminal (i.e. $f\textsubscript{1}, f\textsubscript{2}, f\textsubscript{3}, .. f\textsubscript{n}$). The overall features help analyse how the payment system as a whole is performing. For every transaction, time-window-based features on $t$ seconds and event-based features over $e$ events are calculated by the time-decay method and sliding mean method respectively as explained above. For the 1st payment transaction, the overall feature values corresponding to the time-window-based and the event-based features are equal to $1$.

\subsection{Feature selection}
\label{feature selection}

The performance of a terminal is extremely volatile at the minute level varying throughout the day. Therefore, it is necessary to select the features that capture this volatility. Hence, the importance of these features is correlated to how accurately they represent the ongoing volatile performance of the terminal.

Feature selection is a technique of reducing the number of input variables while developing a predictive model \cite{feature_selection}. Choosing an optimal subset of the features can improve the accuracy of the model, reduce model complexity, and enable faster training times for the algorithm in general.
In this section, the strategy to select the most important features from time-window-based features, event-based features, and overall features that are accurate in tracking the terminal’s performance is discussed.

For feature selection, a two-step feature reduction approach was employed. The first step involves feature reduction using Recursive Feature Elimination (RFE) \cite{RFE_RF} and the second step involves calculating Variance Inflation Factor (VIF) \cite{VIF} values for features and iteratively eliminating high VIF features one by one. RFE works by searching for a small subset of features amongst all features in the training dataset and successfully removing features until the desired number remains. This is accomplished by fitting the random forest learning algorithm to all the features and ranking them by their importance, discarding the least important features, and re-fitting the model. This process gets repeated until a specified number of features remain. Variance inflation factor determines the strength of correlation between independent variables. It is calculated by regressing a variable against every other independent variable. A VIF value greater than $5$ implies the variable is strongly correlated with other independent variables and hence can be removed from training. This process continues until no feature remains with VIF greater than $5$. This ensures that the final subset of features have very low multicollinearity between them and are truly independent of each other.
\begin{figure}[t]
\begin{center}
   \includegraphics[width=1.0\linewidth]{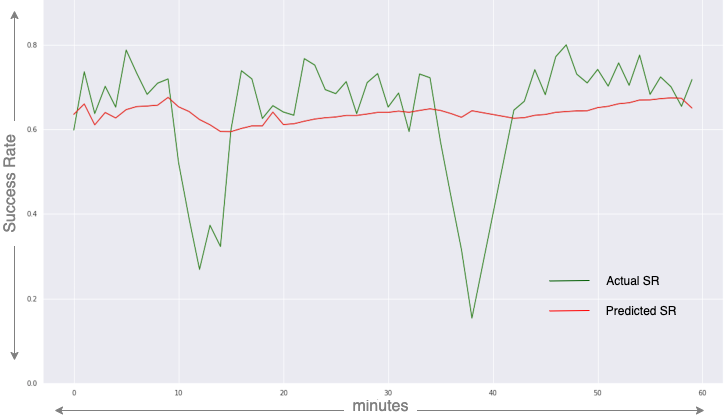}
\end{center}
   \caption{Actual vs Predicted SR using a feature that represents past 2 hours performance (y-axis represents real-time success rate).}
   \label{2hr_feature}
\end{figure}

\begin{figure}[t]
\begin{center}
   \includegraphics[width=1.0\linewidth]{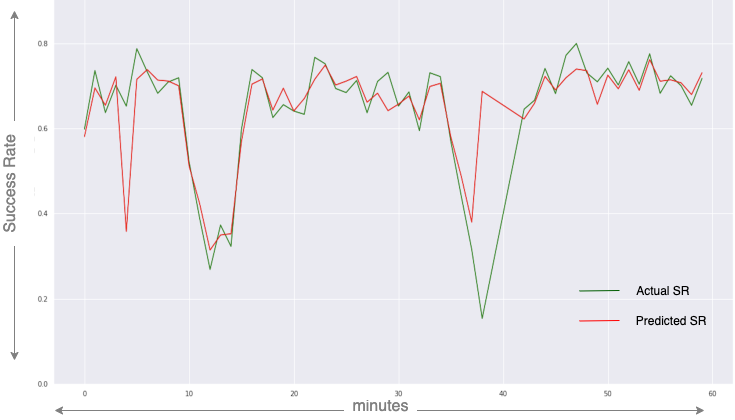}
\end{center}
   \caption{Actual vs Predicted SR using a feature that represents past 5 seconds performance (y-axis represents real-time success rate).}
   \label{5sec_feature}
\end{figure}

\begin{figure}[t]
\begin{center}
   \includegraphics[width=1.0\linewidth]{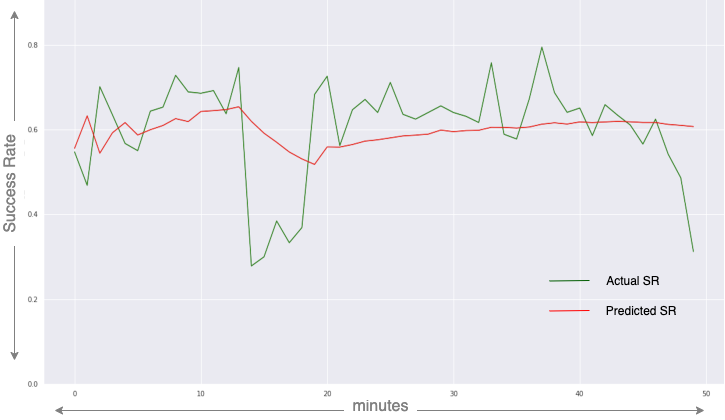}
\end{center}
   \caption{Actual vs Predicted SR using a feature that represents past 300 event performance (y-axis represents real-time success rate).}
   \label{300e feature}
\end{figure}

\begin{figure}[t]
\begin{center}
   \includegraphics[width=1.0\linewidth]{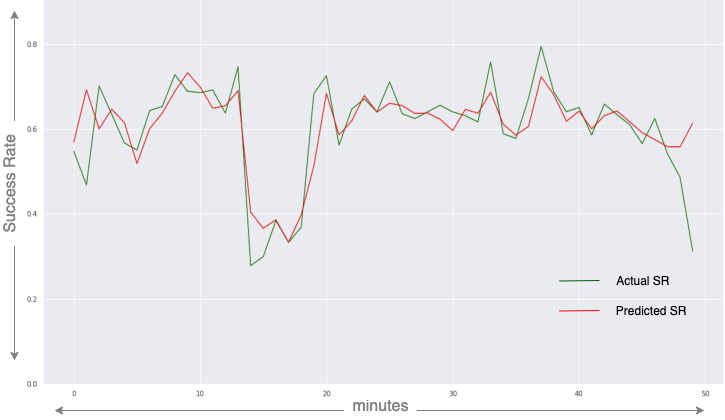}
\end{center}
   \caption{Actual vs Predicted SR using a feature that represents past 10 events performance (y-axis represents real-time success rate).}
   \label{10e feature}
\end{figure}

Feature importance for event-based and time-window-based features is calculated by finding the node impurity \cite{impurity}.
This helped in deducing that feature values with longer time windows (for instance, $2hr$, $3hr$, etc) and large event windows (for example, $200 events$, $300 events$, etc.)  were not as significant as the feature values that depict smaller windows (for example, $5sec$, $30sec$, $10events$, $30 events$, etc). Hence these feature values were excluded from model training.
Fig. \ref{2hr_feature} and Fig. \ref{300e feature} depict that $2hr$ window feature and $300 events$ window feature remain stagnant, while Fig. \ref{5sec_feature} and Fig. \ref{10e feature} show that $5sec$ window feature and $10 events$ window feature capture the terminal volatility and represent the real-time performance of the terminal more precisely.



\subsection{Feature Decay}
\label{feature decay}
\subsubsection{Decay for Feature Creation}

Traditional adaptive systems generally utilise sliding windows over the latest events. In the proposed methodology, the importance of each feature is modified in real-time using a half-life decaying function. Based on the temporal information available about each transaction processed, the feature values are weighted according to its recency. The resulting feature values for the terminal are based on all the past transactions of the terminal, correlating more to the outcome of the recent transactions and less to the older ones.

For example, let feature $f_1\_5s$ be a time-window-based feature representing a terminal's performance for the last $5$ seconds. $f_1$’s significance will reduce to half at $5$ seconds. Here, $5$ seconds not only denotes the time frame for which features were created but also indicates the time that needs to be elapsed for the feature value to obtain half of its initial significance.
Let’s assume that a feature $f$ (from the features list obtained after feature selection) was last updated at $t_1$ and let the current time be $t_2$. Then the updated feature value at $t_2$ will be denoted as shown in equation \ref{decayequation}, where $hl$ is the half-life of the feature in seconds for time-window-based features.

\begin{equation} 
\label{decayequation}
\begin{split}
f\textsubscript{t\textsubscript{2}} = \frac{f\textsubscript{t\textsubscript{1}}}{2 ^ {(t\textsubscript{2} - t\textsubscript{1}) / hl}}
\end{split}
\end{equation}

\subsubsection{Feedback Loop}
\label{feedback loop}
The main purpose of the feedback loop is to track and update the feature values for the time-window-based, event-based, and overall features for a terminal $k$. A low value of the features reflects a low performance of the terminal and vice-versa.
As shown in Fig. \ref{Proposed method}, based on the outcome (success/failure) of the payment processed through a terminal $k$, the selected features corresponding to that terminal will be updated using the adaptive time-decay algorithm shown in equation \ref{decayequation}. 
For example, given a terminal $k$, which has recently seen success as an outcome after a large number of failures, the terminal’s feature values will be updated to drive the terminal to gain a higher probability of success. This is done to show that the terminal has started to process the transactions successfully. If the transactions through the terminal $k$ result in failure again, the feature values are decreased correspondingly to reflect that the terminal is not performing well. However, the previous success will still have an impact, although not significantly. That is, instead of assigning equal importance to the past and the recent events, the current methodology assigns lesser importance to the past events by reducing their contribution to the feature values to half at their half-life.

\subsection{Static Module}
\label{static system}
The inputs to the static module are the terminals that need to be assigned to the payment transaction for processing. Some merchants have a few business rules in place where they would want only specific terminals or gateways to be chosen to process their transactions. The static module takes care of filtering terminals according to these business specifications. After business rule filtering, the filtered terminal list is sent to a logistic regression classifier \cite{logistic_regression} module to detect any gateway down times. For every transaction, relevant real-time time-window-based, event-based, and overall gateway features (explained in \ref{Feature Creation}) are fetched from the database which serves as an input to the logistic classification model. Low values for the features of a gateway imply a low success rate and a low success rate corresponds to a gateway downtime. A gateway downtime suggests terminal downtime for the terminals that map to that particular gateway. Hence, if the model predicts downtime for a gateway, the terminals corresponding to that gateway from the input list filtered by the business rules are removed before sending it to the dynamic module.

\subsection{Dynamic Module}
\label{Dynamic System}
The dynamic module is responsible for predicting the success probabilities for each of the input terminals using the random forest model \cite{rf_for_classification} and updating feature values based on the outcome of transactions. Predictions for the incoming payment request are made on these updated features. Razorpay processes millions of payments per day that can be leveraged to identify the inherent incompatibilities between the payment attributes and terminals. For example, debit cards from a bank A can have a success rate of 45\% with terminal $k_1$ and a success rate of 74\% with terminal $k_2$. The model is trained on historical data to predict such patterns. The dynamic module’s goal is also to track the ongoing performance of the terminals precisely in real-time. It uses an adaptive time-decay algorithm where the features created to represent the terminals are being updated in real-time to track their performance for every transaction. As depicted in Section \ref{feedback loop}, this is achieved by updating the features with every payment transaction’s feedback by considering the payment attributes and the transaction’s outcome (success/failure) from the suggested terminal.
\begin{figure}[t]
\begin{center}
   \includegraphics[width=1.0\linewidth]{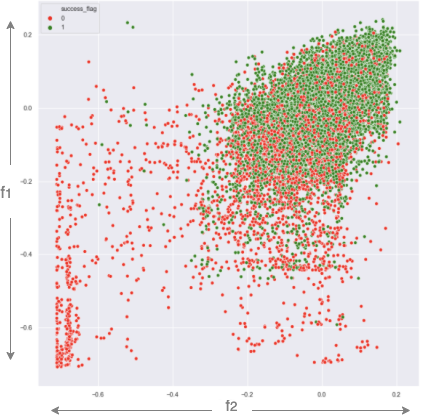}
\end{center}
   \caption{Actual distribution of payment outcomes using top 2 features.}
   \label{actual}
\end{figure}

\begin{figure}[t]
\begin{center}

   \includegraphics[width=1.0\linewidth]{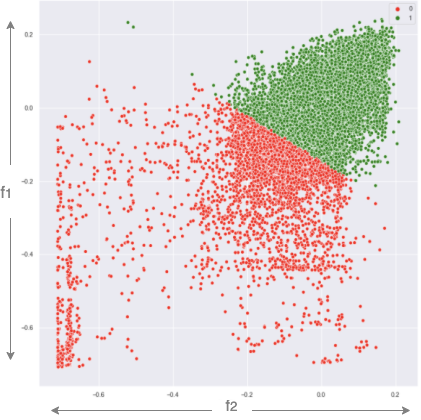}
\end{center}
   \caption{Prediction through Logistic Regression using top 2 features.}
   \label{LR}
\end{figure}

\begin{figure}[t]
\begin{center}
   \includegraphics[width=1.0\linewidth]{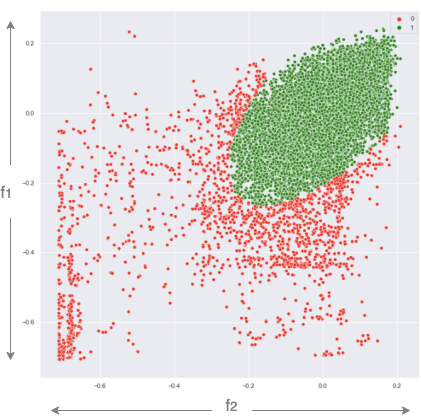}
\end{center}
   \caption{Prediction through Random Forest using top 2 features.}
   \label{RF}
\end{figure}

\par
\subsubsection{Training a classifier for success probability calculation of a terminal}
The probability of success for each terminal is calculated using an optimal subset of features that accurately resemble the performance of the terminals. The feature values created in Section \ref{Feature Creation} and selected in Section \ref{feature selection} act as the independent variables in training the random forest model, and the actual outcome of the transaction (success:1/failure:0) acts as the dependent variable.
Random forest is a bagging-based ensemble method that also gives the probability associated with each target class (here success/failure). The random forest model was trained on approximately $35$ million transactions (excluding transactions with customer-related failures, as terminal/gateway performance does not account for them). The algorithm was tuned to maximise precision for success (explained in Section \ref{results}). Optimal hyperparameters were chosen using the grid-search cross-validation method. This model is trained every week on recent transactions and redeployed to capture latest trends.
Fig. \ref{actual} shows the actual distribution of success(green) and failure(red) for payments plotted against the top 2 features (based on their feature importance). Fig. \ref{LR} and Fig. \ref{RF} show the predictions for the same set of payments and features through a logistic regression model and a random forest model respectively. These figures clearly indicate that the random forest model provides better separation between the outcomes, hence a non-linear tree-based ensemble classifier algorithm was chosen over a linear classifier algorithm for this data.

\section{Results and Impact}
\label{results}
To get the best performing model for the dynamic module, various machine learning algorithms like linear models, tree-based ensemble models, neural networks, etc have been examined. Since the dynamic module needs to return a list of terminals sorted by their success probabilities, it makes business sense that the module does not end up assigning high probabilities to low-performing terminals. In other words, the model should have fewer false positives. Thus, precision has been used as the primary metric to evaluate the model's performance. In this analysis, precision (equation \ref{precision}) and ROC-AUC score (equation \ref{roc_auc}) have been computed for each model. The comparison of these results in Table \ref{Model Comparision} clearly shows that the random forest model outperforms the other models in the desired metric (precision).
\begin{figure}[t]
\begin{center}
   \includegraphics[width=1.0\linewidth]{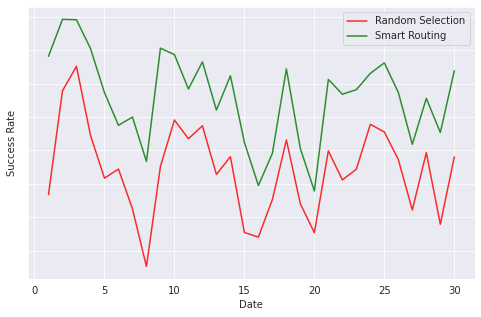}
\end{center}
   \caption{SR comparison of Random Selection vs Smart Routing.}
   \label{SR_comparision}
\end{figure}
\begin{equation} 
\label{precision}
Precision = \frac{TP}{TP+FP}
\end{equation}
\begin{equation} 
\label{roc_auc}
ROC\text{-}AUC = \text{Area under the } TPR \text{ vs } FPR \text{ curve}
\end{equation}

where,
\begin{displaymath}
\begin{aligned}
TP &= \text{True Positive,} \\ 
FP &= \text{False Positive,} \\
TPR &= \text{True Positive Rate,} \\
FPR &= \text{False Positive Rate.} \\
\end{aligned}
\end{displaymath}

To check the effectiveness of the system, A/B testing was carried out for about a month, where half of the total traffic was sent through randomly selected terminals, and the other half was sent through the Smart Routing system. Success Rate through both the methods was observed and analysed. From Fig. \ref{SR_comparision} it can be clearly seen that the success rate through the Smart Routing system outperforms the success rate obtained through random selection of terminals. On average, an increase of 5\% was observed in the success rate through the Smart Routing system.

Being a payment aggregator Razorpay needs to offer an available, credible, transparent, and reliable service to the merchants. They also need to ensure that the merchants can offer a good user experience at optimal costs to the customers using these services.

The Smart Routing solution presented in this paper improves the efficiency of the overall lifecycle of a payment system. It impacts businesses greatly by boosting the success rates thereby increasing their revenues.

Even though the Smart Routing system improves the SR, a terminal might fail unexpectedly during a payment transaction. In such cases, 3 courses of action can be taken: (i) retry from the same terminal again, (ii) retry from a random terminal, (iii) send the request to the next best terminal already predicted by the dynamic module. 
Since there is no information about the success probability of the terminal that failed in the last transaction or a terminal that is randomly chosen, selecting the next best terminal increases the success probability for a retry. This reduces the overall number of retries, which improves the user experience and prevents the wastage of infrastructure resources associated with a payment transaction. Moreover, with the sorted list of best terminals already available, the next best terminal is chosen instantly without sending another request to the prediction module. 

Downtimes are inevitable in any system. The resilience of a system to such downtimes affects the user experience, which in turn affects the business revenues. Merchants and customers gauge the resilience of a system by how quickly it can detect and mitigate such downtimes. The static module helps in identifying the downtime for a gateway and assists in diagnosing the reason for such downtimes. This helps Razorpay filter out terminals associated with the failed gateway before sending the list of terminals to the dynamic module for prediction. This is also used to alert the merchants promptly so that they can take necessary actions to evade business losses.

\begin{table}[]
\begin{center}
\caption{Model comparison and metrics.}
\begin{tabular}{|l|c|c|c|}
\hline
\backslashbox[40mm]{\textbf{Model }}{\textbf{Metrics}}               & \multicolumn{1}{l|}{\textbf{Precision}}  & \multicolumn{1}{l|}{\textbf{ROC-AUC}} \\ \hline
\textbf {Random Forest}       & \textbf{0.9469}                                                 & \textbf {0.7949}                         \\ \hline
XGBoost             & 0.8431                                                & 0.6993                         \\ \hline
Light-GBM           & 0.9433                                                & 0.7130                          \\ \hline
Logistic Regression & 0.9465                                                & 0.7340                         \\ \hline
Neural Networks     & 0.9105                                                & 0.7381                         \\ \hline
\end{tabular}

\label{Model Comparision}
\end{center}
\end{table}

\section{Conclusion}
The Smart Routing solution for payment transactions presented in this paper processes millions of transactions in real-time and provides significant improvements in the success rate for payments. As described, this solution is a pipeline that consists of a static module and a dynamic module. The static module is based on rules and simple ML techniques to prune the list of probable terminals for a given payment transaction. This module helps in exerting fine control over the payment flow by filtering out the irrelevant and poor-performing terminals before sending their data to the dynamic module. The dynamic module uses hand-crafted and dynamically updated features to predict the probability of success for every terminal. These features not only encapsulate the past performance of the terminal but also utilise the impact of other payment attributes while routing the payments. 

This pipeline is highly explainable because of the interpretable nature of the ML models used. This helps in identifying and eliminating the causes for failures, in turn making the payment systems secure against performance dips. In conclusion, this work shows how interpretable ML systems integrate seamlessly with the existing architecture and improve business performance. The routing concepts presented in this work can be reused for various industrial applications where real-time feature updates based on current outcomes affect subsequent predictions. A few notable use cases include stock market prediction, online product recommendations, estimation of advertisement click-through rates, anomaly detection, etc.

\section{Future Work}
The existing results are promising by utilising the supervised learning techniques for Smart Routing of payment transactions. More sophisticated ML algorithms like sequence models can be explored to predict the best terminals for a given payment. 
The current system does not give an opportunity to poor-performing terminals after a certain amount of inactivity, thus the system may miss out on some terminals if their performance is improved within this time span. Building a terminal revival module might help in rejuvenating these dormant terminals more frequently. Additionally, employing better pricing features in the dynamic module may assist in optimizing the cost of payment transactions. 

Feedback systems implemented using reinforcement learning usually improve the efficacy of ML systems. Further work needs to be done to evaluate if utilising reinforcement learning in this pipeline improves the performance of the system. 

{
\small
\bibliographystyle{ieeetr}
\bibliography{egbib}
}
\end{document}